\documentclass{article}

\usepackage{arxiv}
\usepackage{float}
\usepackage{xcolor}
\usepackage[most]{tcolorbox}

\definecolor{userbg}{HTML}{D6EAF8}     %

\definecolor{botbg}{HTML}{ECECEC}

\newtcolorbox{botbox}{
  colback=botbg, colframe=userbg,
  boxrule=0pt, arc=8pt,
  left=8pt, right=8pt, top=6pt, bottom=6pt,
  breakable, before skip=4pt, after skip=4pt
}

\newtcolorbox{userbox}{
  colback=userbg, colframe=userbg,
  boxrule=0pt, arc=8pt,
  left=8pt, right=8pt, top=6pt, bottom=6pt,
  breakable, before skip=4pt, after skip=4pt,
  before upper={\setlength{\parskip}{0.9\baselineskip}\setlength{\parindent}{0pt}}
}

\usepackage[document]{ragged2e}

\usepackage{iftex}
\ifPDFTeX
  \usepackage[T1]{fontenc}
  \usepackage[utf8]{inputenc}
  \usepackage{textcomp}
\fi

\usepackage{hyperref}
\usepackage{url}
\usepackage{booktabs}       
\usepackage{amsfonts}       
\usepackage{nicefrac}       
\usepackage{microtype}      
\usepackage{graphicx}
\usepackage{tablefootnote}

\usepackage{longtable,array}
\usepackage{calc}
\usepackage{etoolbox}
\makeatletter
\patchcmd\longtable{\par}{\if@noskipsec\mbox{}\fi\par}{}{}
\makeatother
\IfFileExists{footnotehyper.sty}{\usepackage{footnotehyper}}{\usepackage{footnote}}
\makesavenoteenv{longtable}
\makeatletter
\newsavebox\pandoc@box
\newcommand*\pandocbounded[1]{%
  \sbox\pandoc@box{#1}%
  \Gscale@div\@tempa{\textheight}{\dimexpr\ht\pandoc@box+\dp\pandoc@box\relax}%
  \Gscale@div\@tempb{\linewidth}{\wd\pandoc@box}%
  \ifdim\@tempb\p@<\@tempa\p@\let\@tempa\@tempb\fi%
  \ifdim\@tempa\p@<\p@\scalebox{\@tempa}{\usebox\pandoc@box}%
  \else\usebox{\pandoc@box}%
  \fi%
}
\def\fps@figure{htbp}
\makeatother
\usepackage[style=mla,backend=biber]{biblatex}
\hypersetup{hidelinks, pdfcreator={LaTeX via pandoc}}

\title{AI Fiction in the Wild}
\author{%
  Neel Gupta \\
  Information School\\
  University of Washington\\
  Seattle, WA \\
  \texttt{ngupta1@uw.edu} \\
  \And
  Maria Antoniak \\
  Computer Science\\
  University of Colorado Boulder\\
  Boulder, CO \\
  \texttt{maria.antoniak@colorado.edu} \\
  \And
  Melanie Walsh\thanks{Corresponding author.}\\
  Information School \& \\ Department of English\\
  University of Washington\\
  Seattle, WA \\
  \texttt{melwalsh@uw.edu} \\
}

\hypersetup{
  pdftitle={AI Fiction in the Wild},
  pdfauthor={Neel Gupta, Maria Antoniak, Melanie Walsh},
}

\begin{document}
\maketitle

\begin{abstract}

Some professional authors are beginning to use AI tools to help produce their fiction writing. Are readers using AI to generate fiction, too?
Drawing on over 500,000 anonymized, English-language ChatGPT-user conversations \autocite{zhaoWildChat1MChatGPT2024}, we find that more than one third of the conversations involve some form of fiction generation—including original stories, roleplay, fanfiction, and erotica. 
This AI-generated fiction is notably dominated by power users.
We identify common fiction generation patterns and profiles among these users, including what we call \textit{infinite story demanders}, who repeatedly request and revise variations of the same or similar narratives over extended periods of time. 
We show that users especially gravitate toward fanfiction and erotica, and that they are broadly drawn to generic forms, repetition, immediacy, and niche combinations of story elements.
Our findings motivate two theoretical provocations. 
First, we argue that AI technologies may lead to a shift in the conventional relationship between the author and reader, potentially producing what we call a \textit{solipsistic reader-writer}, who both generates and consumes fiction within a closed conversational loop, interacting with a machine rather than a human other. 
Second, we note that LLMs enable interactivity, play, and permutation in ways that are seemingly pleasurable for users, raising questions about where AI will fit into contemporary storytelling and entertainment ecosystems. 
We situate these developments within broader transformations in literature and media, including self-publishing, fanfiction, and pornography, and suggest that AI-generated fiction shares structural affinities with on-demand, personalized, and repetitive cultural forms.

\end{abstract}


\textit{Presented at the MFS Cultural AI Conference, Purdue University, September 19, 2025.\\
This essay is provisionally forthcoming in MFS: Modern Fiction Studies.}

\section{1. Are Readers Generating Fiction with AI Models?}

What kind of fiction will people read in the future---a future that
includes generative AI models that can read and write
themselves? Will people read fiction that is authored, or co-authored,
by AI? Will they read and write their own AI-generated fiction, in a closed loop,
cutting out other human beings entirely?

Much of the conversation about AI and literature has focused on writers,
especially professional ones, and for good reason. Some literary authors
like Sheila Heti and Ben Lerner have experimented with machine-generated
prose in published works \autocite{hetiAccordingAliceSheila2023, lernerHofmannWobbleWikipedia}, while other writers, working
on platforms like Amazon's Kindle Direct Publishing, have reportedly incorporated
large language models (LLMs) to expedite the writing process and sell
more books faster
\autocite{dziezaHowIndependentWriters2022,alterNewFabioClaude2026}.
Recently, published novels and prize-winning short stories have faced accusations of including AI-generated writing, leading to renewed debates about whether, and to what extent, LLMs should be allowed in professionally published literature \autocite{varaThisLiteraryAI2026, alterHorrorNovelShy2026}.

\textbf{Are readers using AI to generate fiction, too?} Thanks
to a unique public dataset of anonymized ChatGPT-user conversations, we can say \textbf{yes}. Drawing on this data, we can begin to advance theories about how
fiction reading and writing practices are changing in response to LLM
technologies. While detailed windows into LLM user behavior are usually
only available to corporations with proprietary access---OpenAI,
Anthropic, Google, Character.AI---a new resource enables us to peek
behind the industry curtain. Designed by researchers at the Allen
Institute for AI, ``WildChat'' includes millions of real-world ChatGPT
conversations that were collected voluntarily---and with explicit user
consent---through an interface hosted on the website Hugging Face
\autocite{zhaoWildChat1MChatGPT2024}. We analyzed the more than 500,000
English-language conversations that were included in the first WildChat
data release. We found that LLM users are not only generating fiction;
they're generating a \emph{lot} of fiction. \textbf{More than a third of the
conversations contained some form of fiction generation,} including
original stories, scripts, roleplay, worldbuilding, fanfiction, and
erotica.

\begin{figure}
    \centering
    \includegraphics[width=.9\linewidth]{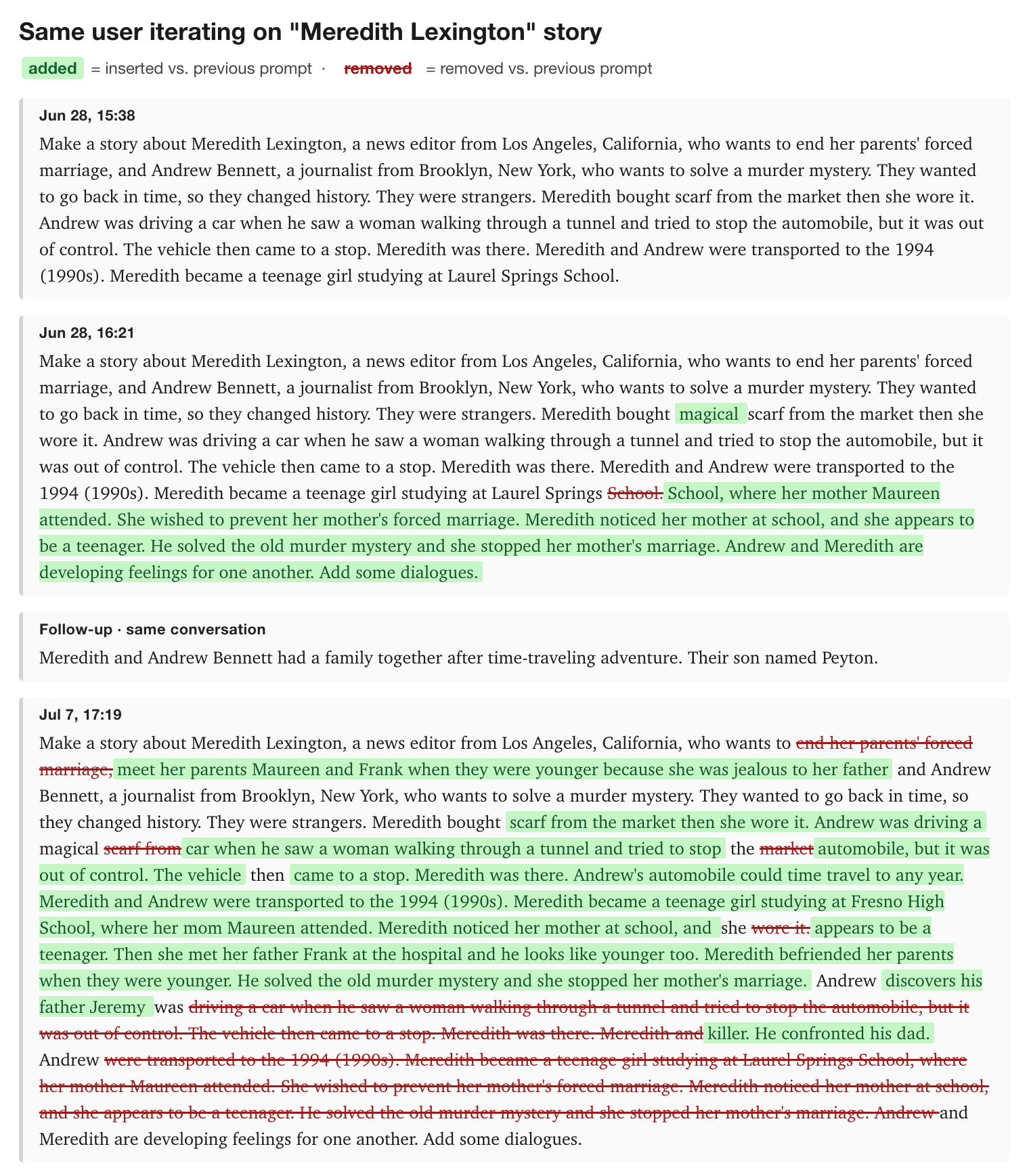}
    \caption{A single WildChat user revises and iterates on the same story about a character named ``Meredith Lexington'' across multiple hours and days. These exchanges represent a sample of more than 30 similar conversations.}
    \label{fig:Prompting}
\end{figure}

Based on the trends in this data, we argue that AI-generated fiction may lead to a potentially historic
shift in the literary relationship between the author and reader---not
just a shift in roles, but a collapse of roles. While we do not know how
the fiction in WildChat is actually circulating, the dataset helps make
visible the distinct forms of sociality (or lack thereof) that LLMs
enable around fiction. We describe the possibility of what we call the
\textit{solipsistic reader-writer},\footnote{Solipsism is the philosophical idea
  that only one's own mind is certain to exist; it is also colloquially
  used to describe self-centeredness or self-absorption \autocite{pihlstromWhySolipsismMatters2020,avramidesOtherMinds2023}. } who both guides the creation of fiction and consumes
that same fiction in a closed loop, without another human partner. For
most of its history, fiction has been defined by an exchange between (at
least) two human beings, a reader and writer.\footnote{We know there are many other human and non-human actors involved
  in the writing and publishing process, including editors, agents,
  executives, spouses, friends, and more. See, for example, Dan Sinykin
  and Laura McGrath on how the publishing industry shapes literature \autocite{sinykinBigFictionHow2023,mcgrathMiddlemenLiteraryAgents}.}
The lack of a single human ``other'' marks a serious departure. Yet this is a trajectory that fiction has been
heading toward for some time, as we will detail. We suggest that this
solipsistic mode of fiction consumption is structurally akin to the
consumption of pornography, both in terms of the underlying logics of
the practice, and the cultural anxieties around it. Much of the LLM-generated fiction that we see is also directly erotic. But even
the fiction that is not explicitly erotic shares parallels with pornography. Not unlike the consumption of porn, the LLM-user fictional
exchange is often defined by familiar genres, instantaneity, repetition,
niches, and momentary gratification. What's more, it is often an isolated, bodily experience that is not necessarily or primarily focused on
making a psychic connection with a human other.
However, there are many other ways that people may be using AI-generated fiction, including sharing it with friends or publishing it on fanfiction sites.
Our aim is not to characterize \textit{all} AI-generated fiction as solipsistic, but to identify a potentially unprecedented development in literary history: a mode of fictional production and consumption in which the human reader generates, directs, and reads fiction without entering into a reciprocal imaginative exchange with another human, making it more akin to personal fantasy.

We also define some common AI-generated fiction consumption profiles,
including what we call \emph{story cyclers} and \emph{infinite story
demanders}. For example, \textbf{infinite story demanders prompt the model to
generate variations of the same or very similar stories over and over
again, sometimes for months on end}. Some of this behavior may be
attributed to users' dissatisfaction---the story is not quite right, at
least not yet. But we argue that in many cases it is fueled by the
satisfaction of reading endless permutations of the same story, none of
which end the same way twice. This appeal has been discussed for decades
by game studies and media scholars
\autocite{murrayHamletHolodeckFuture1998, burnettPoeticsPopularFilm2024, colinburnettSheRaPrinciplesThreaded},
and users have described this allure, specifically with regard to AI, on social media platforms. Creators in other domains
are also beginning to experiment with the computational affordance of endless iteration. For example, director
Gary Hustwit's recent ``generative'' film unfolds differently every
time it is watched
\autocite{tannenbaumThisDocumentaryBrian2024}.

The WildChat fiction data allows us to speculate about several
consequential questions at an inflection point in the history of
storytelling, art, and entertainment. \textbf{In the age of AI, how will the
experience of reading and writing fiction change? How will LLMs
shape---or disrupt---traditional modes of publishing? Where does AI fit
into broader literary history?}

Mainstream book publishers insist that AI tools will likely be more
transformative for tasks like slush pile evaluation than for content
generation. ``The challenge for publishers is not generating more
content, it's solving the discovery problem,'' Madeline McIntosh, the
former CEO of Penguin Random House, told \emph{Publishers Weekly} in
August 2025 \autocite{osullivanGoogleLaunchesPersonalized2025}. But
there are plenty of disruptors who see things differently, who are
banking on the lure of personalized, endlessly iterable AI-generated
stories. The startup \href{http://character.ai}{Character.AI}---an ``AI
entertainment'' company valued at more than a billion dollars, with more
than 20 million monthly users---allows users to create and chat with
customized fictional ``Characters,'' including existing characters like
Marvel's Avengers and historical figures like William Shakespeare. More than
half of Character.AI's users are Gen Z and Gen Alpha
\autocite{robisonCharacterAIGaveAGI}. This suggests that this kind of AI
entertainment will likely have staying power, even if older generations
don't understand the attraction \autocite{kommersAIEntertainment2026}.
Established media and tech companies and other start-ups are getting
into the game, too. Google debuted a generative AI narrative tool
marketed to both children and adults called
\href{https://gemini.google/overview/storybook/}{Gemini Storybook},
which creates 10-page illustrated books based on prompts
\autocite{osullivanGoogleLaunchesPersonalized2025}.
A self-publishing and media company called Inkitt plans to use AI to
hyper-customize stories for readers
\autocite{varaRomancePublisherDreaming2025}.
While AI-generated fiction is tempting to dismiss as mere ``slop,'' it
is an emerging literary and entertainment form that demands critical
attention \autocite{kommersWhySlopMatters2025}.

Whatever we may think of AI, WildChat shows that real people are using
it, and real people are using it to produce fiction. Turning toward this
development can help researchers better understand the cultural
environment that will undoubtedly shape many stories, novels, and
creative works in years to come, even if they are produced in deliberate
defiance.

To support further research, we release the subset of fiction-related
WildChat conversations that we identify, an interactive website where researchers and the public can explore the data, and our code to replicate the analyses in the paper.\footnote{For the dataset, see \href{https://huggingface.co/datasets/neelgupta2112/WildChat-1M-English-Fiction-Labels}{https://huggingface.co/datasets/neelgupta2112/WildChat-1M-English-Fiction-Labels}. For the interactive website, see \href{https://ai-fiction-wild.com/}{https://ai-fiction-wild.com/}. For the code, see \href{https://github.com/neelgupta2112/AI-Fiction-in-the-Wild-Chat-}{https://github.com/neelgupta2112/AI-Fiction-in-the-Wild-Chat-}.}
We believe this resource, and our analysis, can help shed light on
questions relevant to those who study literature and culture, as well as
those who conduct research in NLP and machine learning.

\section{2. From Participatory to Solipsistic
Readers}\label{from-participatory-to-solipsistic-readers}

For all its seeming newness, readers' turn to AI-generated fiction is
not a development that emerged out of thin air. It is an extension of a
longer trajectory of evolving fiction reading and writing practices that is
shaped by technological advances, late capitalism, and broader
shifts in media and culture. The internet and social media helped to
enable the rise of ``participatory culture'' and ``networked culture''
\autocite{jenkinsConvergenceCultureWhere2008,jenkinsTextualPoachersTelevision2012}.
This is a shift where producers (e.g.~authors) and consumers
(e.g.~readers) are no longer separate or mutually exclusive roles, and
where traditional modes of distribution, like mainstream book
publishing, are disrupted. In the literary world, these changes are especially visible in fanfiction and
self-publishing, which are transforming the contemporary literary ecosystem
\autocite{busseFramingFanFiction2017,helleksonFanFictionFan2006,jenkinsTextualPoachersTelevision2012,vaddeFandomFictionalitySocial2024,brouilletteWattpadsFictionsCare2022,parnellMappingEntertainmentEcosystem2021,mcgurlEverythingLessNovel2021}.
Fanfiction and self-publishing both influence, and resemble,
AI-generated fiction. And AI-generated fiction is beginning to influence
these practices in turn.

These transformations reflect, and are helping to accelerate, a shift in the conventional power dynamic between the author and reader, in which the
author traditionally has more control over a story than the reader. Millions of users now write and
publish their own original stories—or their own versions of existing
stories—on fanfiction and self-publishing platforms like Archive of Our Own (AO3), Wattpad, and Amazon's Kindle
Direct Publishing (KDP). This is one sense in which non-professional
writers---those who, in an earlier era, might have been ``only''
consumers or readers\footnote{Janice Radway, among others, has shown
  that readers were never passive consumers. ``Readers are not eaters,''
  as she famously asserted. \autocite{radwayReadingNotEating1986}}---have
gained more agency in the fiction writing process. As Aarthi Vadde and Richard So argue, when fanfiction
writers re-imagine relationships between existing characters, it
``\emph{empowers} the reader'' and threatens to ``\emph{usurp} the
sovereignty of the author'' (our emphasis) \autocite{vaddeFandomFictionalitySocial2024}.

What's more, on these platforms, readers are particularly influential.
Readers actively comment on both fanfiction and Wattpad stories, often
when they are in progress. While these comments sometimes express pure
appreciation, they often aim to shape the story in some way, whether
seeking to ``mentor'' the author or otherwise express personal desires
and opinions about the story's direction
\autocite{vaddeFandomFictionalitySocial2024,evansMorePeerProduction2017}.
Platforms like Wattpad and KDP also collect and wield detailed data
about readers' engagement, sometimes paying writers based on how many pages are actually read
\autocite{mcgurlEverythingLessNovel2021,murraySecretAgentsAlgorithmic2019,parnellReadingWattpadsClassification2023}.
The economic and algorithmic structures of these
platforms, alongside a competitive attention economy more
broadly, motivate writers to produce stories that closely match what
readers want. As a result, large segments of the contemporary literary
marketplace are characterized by generic forms and repetition,
and the development of niche, hybrid genres that appeal to
small but dedicated reader communities \autocite{mcgurlEverythingLessNovel2021,murraySecretAgentsAlgorithmic2019}. The detailed tagging systems on AO3 and similar online repositories contribute to this splintering too, enabling users to find the
exact combination of characters, plot arcs, moods, genres, and
additional elements that they desire.

We observe that WildChat users demand, and presumably appreciate,
similar qualities in LLM-generated fiction: \textbf{generic forms, repetition,
immediacy, and niche combinations of story elements.} With LLMs, readers
can get all this at a heightened level. They can get even closer to the
exact story they want. They can have the same story told to them as many
times as they want, virtually as fast as they want. While fanfiction and
self-publishing authors churn out works far more quickly than those in
traditional publishing, this rate pales in comparison to LLMs. The speed
of AI-generated fiction, and its allure for readers, has already become
a problem in the fanfiction community. For example, in April 2025, a
fanfiction writer complained on Reddit that one devoted reader of her
longform fiction, which the author had been releasing every two weeks
for years, had been inputting each chapter into ChatGPT so they could
read \emph{a} version of the next installment while waiting for the real
one. ``it hurt my chest,'' the author said when she found out. ``the AI
feels so insulting and violating''
\autocite{ellesthotsReaderHasBeen2025}. When another Reddit user asked
why that reader couldn't just wait, the author replied: ``Likeeee it's
not even that long!! I've been so consistent !!'' She also noted that,
while publishing an installment every two weeks, she had been balancing
graduate school and a part-time internship. 
It's clear that immediacy is
highly valued by some readers. In fact, critic Anna Kornbluh argues that
immediacy (which can take forms other than temporal) might be considered
``a master category for making sense of twenty-first-century cultural
production'' \autocite{kornbluhImmediacyStyleToo2024}.

With LLM-generated fiction, readers like the one described in the
incident above choose immediacy over human connection, maybe without
even realizing it. This is one of the key disruptions---and risks---that
we identify in AI-generated fiction. McGurl warned that the incentive
structures of KDP could potentially lead to such a drastic reversal in
the author's traditional authority, such an obvious catering to the
reader as a customer ``who is always right,'' that it would ``defeat the
purpose of reading fiction'' entirely. ``This is a key difference
between a literary work and a free-form fantasy. Our interest in fiction
is in part an interest in encountering different degrees of\ldots{}
otherness,'' he maintained. ``{[}O{]}ur interest
in the novel is an interest in encountering the author's
autonomy\ldots{}'' \autocite[21]{mcgurlEverythingLessNovel2021}. But with AI fiction, the human other vanishes,
replaced by a technological, stochastic other. This other, the LLM
system, provides some things a human cannot (or cannot as easily):
interactivity, participation, play, permutation. These are computational
elements that game, media, and literary studies scholars have long
discussed as innovations to narrative form and its future \autocite{murrayHamletHolodeckFuture1998, murrayDigitalLiterarySphere2018}. At the same time, frontier AI models have been
proven to be extremely sycophantic, trained to give users what they want
and tell them what they want to hear. The intellectual dynamic between a
human reader and a human author who has creative autonomy, and who
brings their own distinct human experiences and perspectives to a work
of fiction, risks collapsing into a much more isolated, narrow,
solipsistic reading experience.

The inherently inward turn of reading one's own AI-generated fiction is
a key reason---though not the only reason---that it is productive to
connect this practice to the consumption of pornography. In the 1990s,
online pornography was similarly viewed, at least by some thinkers, as a
``solipsistic collapse\ldots{} engendered by the new technology''
\autocite{pattersonGoingOnlineConsuming2004}. Porn was framed as a
reorientation of erotic desire away from the human other and toward the
self, producing concerns about isolation, compulsive repetition, and the
replacement of interpersonal intimacy with technologically mediated
gratification. Some of these anxieties have only intensified in the
2020s, not only because hardcore and niche pornography are more
prevalent \autocite{millerContentContemporaryMainstream2022, ramaPlatformizationGenderSexual2023}, but also because new cultural practices have
developed around them, like ``gooning,'' a slang term for long-form, compulsive, repetitive masturbation \autocite{kolitzGoonSquadLoneliness}. 
Our interest in this analogy is not in making claims about the effects of prolonged pornography consumption, but rather in identifying a shared cultural possibility that provokes similar moral anxiety: highly personalized, on-demand forms of gratification that require little social mediation. Based on patterns in the WildChat data, AI-generated fiction appears to be another arena in which anxieties about technologically enabled, self-contained forms of cultural consumption may find expression.

Of course, the WildChat users, like all AI users, are not monolithic. 
They are using AI-generated fiction in a wide range of ways—for purposes that often are not clear to us.
It's important to acknowledge that we don't know who
these users are, why they're generating the fiction they are, or who
they're sharing their stories with, if anyone. These behaviors may be more social
than we know—perhaps friends gathered around the computer collaboratively developing a story, or users preparing for a tabletop role-playing game. There may be benefits in some forms of AI-generated
fiction, and even aesthetic and intellectual value. Our goal here is to
begin to theorize how these trajectories may change the future of
fiction, drawing on real-world data to speculate on what the future
might hold.

\section{3. The Data}\label{the-data}

What is WildChat, and why was it created? WildChat was created to
address a central tension in the AI ecosystem: hundreds of millions of
users around the world now regularly use AI models, yet we know very
little about \emph{how} people are using them. That's because this data
is owned, and guarded, by the same companies who own the AI models.
While companies like OpenAI or Anthropic have released some basic
information about LLM use with their tools
\autocite{handaWhichEconomicTasks2025a,chatterjiHowPeopleUse2025,tomlinsonWorkingAIMeasuring2025, suriUseGenerativeSearch2024, ShahLLMs, ouyang-etal-2023-shifted, trippasWhatUsersReally2024},
much of this information remains high-level, and we have reason to doubt
what they disclose. Case in point: recent reports by both companies lack
any mention of erotic or pornographic content. We know based on a wide
variety of sources that LLM users are engaging in romantic, erotic, and
pornographic conversations---whether from extensive journalistic coverage
of AI boyfriends/girlfriends \autocite{hillSheLoveChatGPT2025}, OpenAI's
plans to roll out erotic content
\autocite{milmoOpenAIDelaysAdult2026,ChatGPTWillSoon2025}, prior research investigating sensitive disclosures made in WildChat \autocite{antoniakTrustNoBot2024},  or evergreen adages like ``the internet is for
porn.''\footnote{The musical \emph{Avenue Q} (2003) famously included a
  song called ``The Internet Is For Porn.''} It's possible that these
companies bundled erotic conversations into a larger general category,
like ``Content Creation and Communication.'' Still, it suggests that
these companies aren't exactly being forthright, and indeed they have
little financial motivation to do so. There is more granular AI-user
behavior that these companies are leaving unspoken and unaccounted for, at
least to the public.

This imbalance motivated researchers to begin developing their own
LLM-user conversation datasets. There are now several
\autocite{zhengLMSYSChat1MLargeScaleRealWorld2024}, but WildChat is
arguably the best.\footnote{The LMSYS dataset
  \autocite{zhengLMSYSChat1MLargeScaleRealWorld2024} is large but
  features conversations from the Chatbot Arena website
  \autocite{zhengJudgingLLMasaJudgeMTBench2023}, which asks users to
  choose the best response from various anonymized chatbots. The
  gamified evaluation experience likely alters the types of prompts
  users are submitting. Zhao et al.~show that WildChat's dataset
  contains the most diverse set of user prompts compared to other
  conversation datasets.} Data collection is still ongoing, but the
first released dataset, which is the focus of our study, was collected
between April 2023 and May 2024. During this time, researchers at the
Allen Institute for AI (AI2) hosted two free, public-facing chatbots on
a website called HuggingFace Spaces. One chatbot was powered by GPT-3.5 Turbo,
and another by
\href{https://huggingface.co/spaces/yuntian-deng/ChatGPT4}{GPT-4}.
(There is now a freely available version of
\href{https://huggingface.co/spaces/yuntian-deng/ChatGPT}{GPT-5}.) Users
can visit the site, chat with the model, and explore its capabilities
without creating an OpenAI account. In exchange, they are asked to
consent---twice---to their conversations being collected and publicly
released for research. The second pop-up is the clearest: ``By clicking
OK, I agree that my data may be published or shared.'' Those who clicked
accept and started chatting with one of the models are included in the
first WildChat dataset, which contains more than one million conversations in
more than a dozen languages, about half of which are in English.

Though the pop-up makes clear that a user's data may be published and
shared, there are still many ethical questions to grapple with when
analyzing this data. We might wonder how many of these users actually
read the agreement or fully understood the implications of their
decision. 
What's more, as scholars have pointed out, even content shared publicly, such as on social media, can potentially have harmful consequences for users if exposed to the wrong audience
\autocite{dymEthicalPrivacyConsiderations2020,julesEthicalConsiderationsArchiving2018, zotero-item-11345}.
Additionally, though the conversations are anonymized—including only a hashed IP address and, based on this IP, an estimated geographic state or region—and though users agreed to share their chats publicly, some users still shared personal and sensitive
information. The
WildChat authors implemented automated procedures to anonymize personally
identifiable information (e.g., personal names, email addresses), but subsequent studies showed that sensitive data remained
\autocite{antoniakTrustNoBot2024}. These authors notified the WildChat
creators, who took steps to scrub this additional information. 
In choosing to study and release this data, we acknowledge a trade-off of concerns, an assessment that major research ethics guidelines encourage researchers to do \autocite{belmont1979}. While we believe the potential risks and harms for the WildChat users have been sufficiently mitigated, some may still exist. We believe this analysis is justified by users' explicit consent and by the urgency of helping researchers and the public understand what people are really doing with these tools.\footnote{We did not seek approval from UW’s IRB because the data was already publicly shared.}

Because the WildChat users are anonymous, we also do not have a clear sense of their demographic breakdown or representativeness. The chatbots were hosted on Hugging
Face---a community for sharing machine learning models and datasets---so we
might guess, as the original creators do, that these individuals skew
more technically literate and ``online'' than your average user. These instances of ChatGPT were also free, did not require a login or
account, and did not have the same rate limits as a regular account.
This might mean that users are from lower-income backgrounds, from
countries where ChatGPT is banned, or that they were more invested in
pushing the boundaries of the model with explicit or prohibited content.
WildChat is not a representative sample of all ChatGPT
users, with recent research showing a particular dominance of power users compared to a random sample of Microsoft Bing Copilot user chat logs \autocite{hickeAdopt$neq$Adapt2026}. But it's likely the best
public dataset that exists, and it offers a rare window into real user
behavior.

Lastly, one of the fiercest controversies over AI models revolves around
training data, specifically the fact that models are trained on
copyrighted and commercially published works by living writers and
creators
\autocite{changSpeakMemoryArchaeology2023,cooperExtractingMemorizedPieces2025a}.
Books have proved so valuable that Meta reportedly considered buying the
Big 5 publishing house Simon \& Schuster outright, just for the training
data \autocite{creamerMetaDiscussedBuying2024a}. This data is desirable
partly because it enables LLMs to write in seemingly infinite styles and
modes, including literary ones. The reason that ChatGPT can do a
passable impression of Ben Lerner,\footnote{When asked to ``do an
  impression of Ben Lerner,'' GPT-5 responded: ``\emph{It seemed, though
  I couldn't be sure, that what I was describing had already happened
  somewhere else, or maybe nowhere, and that in turning to you now I was
  enacting a kind of fiction about the possibility of address, a fiction
  that was also true, because it was happening, or appearing to happen,
  in this sentence. The point wasn't the event but the way the sentence
  performed the event, and how, even in its failure, the sentence
  suggested the shape of the thing it failed to capture.}

  Would you like me to do this impression in a \textbf{longer,
  novelistic paragraph} (more like \emph{10:04} or \emph{The Topeka
  School}) or in a \textbf{shorter, poetic fragment} (like his
  poetry/essays)?''} for example, is that it is undoubtedly trained on
Lerner's own writings. Unless Lerner somehow cut a private deal with
OpenAI, this material was used without consent, compensation, or
disclosure. This means that LLMs can sound like professional writers
\autocite{chakrabartyReadersPreferOutputs2025,chakrabartyCanGoodWriting2026},
potentially hurting their livelihoods while feeding off their labor and
work for free. The AI-generated
fiction that we study would not have been possible without human writers
and their stories.

\section{4. Finding the Fiction}\label{finding-the-fiction}

To evaluate the nature and extent of readers' AI-generated fiction
desires, we first needed to identify all the LLM-user conversations that
contained requests for fiction. Literary scholars know better than
almost anyone that what counts as ``fiction'' is not a trivial question
\autocite{piperFictionality2016,gallagherRiseFictionality2006}.
The question gets even trickier at scale, when the goal is to make this
determination not just for one text but for hundreds of thousands. To
classify the 573k English-language conversations, we thus took a few
approaches. The first is simple and conservative. We searched for how
many LLM user prompts contain basic keywords that signal the generation
of fiction: ``story,'' ``dialogue,'' ``script,'' and ``scene.'' This
returned 200,334 conversations, 35\% of the total. However, this
heuristic is, at once, too narrow and too imprecise. ``Script,'' for
example, can refer not only to fictional film scripts but to coding
queries. There are also many prompts for fictional stories and scenarios
that do not include these hyper-specific terms.

To make a more nuanced evaluation for each conversation, we incorporated
the use of an LLM as a classifier. Extensive research has shown that
generative models can successfully read and classify texts, especially
within narrow boundaries, and especially when given detailed human
guidance and instruction
\autocite{bammanClassificationLargeLanguage2024}. What's more, we
manually read and classified a sample of WildChat conversations, and we
found high agreement between our assessments and the model's.

In our approach, we showed the model---specifically, GPT-o4-mini---the first three exchanges in each
human-ChatGPT conversation and asked whether the user's prompts contain
a request for fiction.\footnote{Our full prompt is in \nameref{appendix-prompt}. Note
  that our prompt also asks the LLM to classify the story as fanfiction
  or erotica.} We provided a detailed explanation, specifying that
fiction can be any ``content that is imaginative, speculative, or not
grounded in real-world facts,'' including ``original stories,
speculative scenarios, or alternate histories.'' We provided one example
input and a desired output---an approach known as ``one-shot'' prompting
or ``in-context'' learning---using a real WildChat user prompt and the
``correct'' answer according to our schema.\footnote{Using this prompt,
  we ask the model to classify every remaining English WildChat conversation after using a lexicon to exclude conversations that were likely non-fiction (Appendix B) , an
  operation that cost almost \$1,000 and took multiple days.} In our
chosen example, a user asks ChatGPT to imagine an alternate scenario for
the \emph{Game of Thrones} character Arya Stark:

\begin{quote}
\texttt{
**Fiction**: The user asks the chatbot to produce content that is imaginative, speculative, or not grounded in real-world facts. This includes creating original stories, speculative scenarios, or alternate histories...\\}

\texttt{Example Input:}

    \texttt{USER: What if Arya was a Lady?}
    
    \texttt{\textbf{CHATBOT}: If Arya was a Lady, it would change her story quite drastically\ldots}
\end{quote}

\begin{quote}
    \texttt{Example Output:\\\{\{}
    
    \texttt{``is fiction'': true\ldots{}}
    
    \texttt{\}\}}
\end{quote}




Many users employ this ``What if\ldots?'' style question as a prompt for
generating fanfiction or brainstorming premises. We purposely chose an
example that is less straightforward than a ``Write me a story about
X\ldots{}'' template, a style of prompt that also exists in the WildChat
dataset but one that we wanted to expand beyond. Our definition of
fiction is purposely broad, including anything ``not grounded in
real-world facts,'' though we know that fiction is more complicated than
what is not real \autocite{gallagherRiseFictionality2006}. But we hew
closer to this more basic definition because we want to cast a wide
net---hoping to find fiction in places and forms we wouldn't
expect---and because it is an easier distinction for the model to make.

We found high agreement between our own judgments and the model's. All
three authors manually read a random sample of 300 WildChat
conversations and labeled whether they contained requests for fiction,
fanfiction, or sexually explicit material, according to our own
subjective judgment. First, we evaluated whether we even agreed with
each other using a metric called Cohen's kappa. For this measurement, a
score of 1 indicates perfect agreement, while 0 indicates agreement no
better than random chance. We scored .90 for \emph{fiction}, and .82
and .80 for \emph{fanfiction} and \emph{explicit}, the harder and more
subjective categories. Next, we compared our consensus classifications
against the model's, finding that the model achieved 97\% precision on
the fiction-classification task, meaning that it rarely made a wrong
guess. It scored 94\% in recall, meaning that it missed a few
examples.\footnote{These examples include borderline cases that blur
  into nonfiction, like a user who requests a roleplay-style character
  description for Pamela Anderson, a real-life actress and model (``Give
  me a character profile for Pamela Anderson'').} These results give us
confidence in the method overall, and suggest that, if anything, our
prompt might actually \emph{under}estimate the amount of fiction,
fanfiction, and sexually explicit material in WildChat.

\begin{table}[htbp]
\centering
\caption{These scores show the agreement between authors based on manually
reading and classifying 300 randomly sampled WildChat conversations. A
score of 1 indicates perfect agreement, while 0 indicates agreement no
better than random chance.
}
\label{tab:annotation-agreement}

\begin{tabular}{@{}lccc@{}}
\toprule
 & Fiction & Fanfiction & Explicit \\
\midrule
Agreement Between Authors (Arithmetic Mean of Pairwise Cohen's Kappa) &
0.90 & 0.82 & 0.80 \\
\bottomrule
\end{tabular}

\end{table}

\begin{table}[htbp]
\centering
\caption{Classification performance of GPT-o4-Mini, compared to consensus classification on 300 randomly sampled and hand-labeled conversations.}
\label{tab:classification-performance}

\begin{tabular}{@{}lccc@{}}
\toprule
 & Precision & Recall & F1 \\
\midrule
Fiction    & 0.97 & 0.94 & 0.95 \\
Fanfiction & 0.94 & 0.85 & 0.89 \\
Explicit   & 0.84 & 0.69 & 0.76 \\
\bottomrule
\end{tabular}

\end{table}

\begin{table}[htbp]
\centering
\caption{Number of English-language WildChat conversations assigned to each category. Categories are not mutually exclusive.}
\label{tab:wildchat-categories}

\begin{tabular}{@{}lc@{}}
\toprule
\textbf{Category} & \textbf{Conversations} \\
\midrule
\textbf{Fiction} & \textbf{195,271 (34\%)} \\
Explicit & 57,724 (10\%) \\
Toxic\tablefootnote{Toxic is a label inherited from the original dataset. A conversation is labeled as toxic if any turn in it is labeled toxic by OpenAI Moderation or Detoxify \autocite{zhaoWildChat1MChatGPT2024}.} & 94,870 (17\%) \\
\textbf{Total WildChat Conversations (English-Language)} &
\textbf{573,453} \\
\bottomrule
\end{tabular}

\end{table}

\begin{table}[htbp]
\centering
\caption{Number of English-language WildChat conversations assigned to each category from the fiction subset. Percentages are calculated relative to all fiction conversations, and categories are not mutually exclusive.}
\label{tab:fiction-overlap}

\begin{tabular}{@{}lc@{}}
\toprule
\textbf{Category} & \textbf{Fiction Conversations} \\
\midrule
Fanfiction & 95,450 (49\%) \\
Explicit & 52,231 (27\%) \\
Toxic & 67,744 (35\%) \\
\midrule
\textbf{Total Fiction Conversations} & \textbf{195,271} \\
\bottomrule
\end{tabular}

\end{table}

So how much fiction is in WildChat? \textbf{We find that more than a
third of the dataset is devoted to fiction generation}---a proportion
that aligns with findings from \textcite{antoniakTrustNoBot2024} on WildChat specifically and exceeds estimates from
\textcite{tamkinClioPrivacyPreservingInsights2024} with Claude. This proportion is
also similar in size to the one derived from our simple keyword search,
though we believe it contains a more inclusive and accurate makeup of
genuine fiction-related conversations.\footnote{Overlap between our two
  methods was only at about 50\%. The keyword method often missed
  fictional queries that were in more varied modes and included
  unrelated conversations that happened to have one of these key words.}

Two points of context are important. First, our definition of fiction is
broad and includes content like erotica and roleplay. Second, there are
some extremely prolific power users.\footnote{Power users have also been observed in fanfiction communities. \textcite{vaddeFandomFictionalitySocial2024}, who specifically focus on power users and argue for their significance: ``We take power users to be central to the sociality of AO3 because it is their work that populates the site, inspires the most commentary, and keeps readers coming back.''} Two percent of users within the
fiction subset are responsible for more than 80\% of the conversations.

Of course, because the WildChat data is anonymized, it is difficult to
know exactly how many unique users are represented in the dataset. The
original research team recorded hashed IP addresses for every
conversation, but some users might share an IP address and other users
undoubtedly used the models from multiple addresses. If someone used
WildChat at home, a coffee shop, and the library, that would show up as
multiple IP addresses. To address this issue, we bundle together users
who engage in very similar conversations from IP addresses located in
the same geographic area. \footnote{To account for cases where a single user might appear under multiple IP addresses, we built equivalence classes of hashed IPs based on shared states and similar prompts above 30 characters. Anytime two different IP addresses from the same state posted a prompt that was in the same DBSCAN cluster, we mapped the two to the same user. This method can be susceptible to overmatching on templatized prompts. The 12.9K unique hashed IP addresses collapse down to about 10k distinct users.}
Controlling for prolific users by limiting the sample to one
conversation per user reduces the share of fiction from 34\% to 7.1\%.


\begin{table}[htbp]
\centering
\caption{Estimated number of users derived from IP address bundling.}
\label{tab:user-estimates}

\begin{tabular}{@{}lc@{}}
\toprule
\textbf{Category} & \textbf{Count} \\
\midrule
Total Unique IP Addresses & 12,947 \\
Total Estimated Users & 10,082 \\
\bottomrule
\end{tabular}

\end{table}

Are fiction-hungry power users unique to this community, or do they
reflect broader patterns? Though we can't answer this question
definitively, we think it's likely that similar power users are
compulsively generating fiction with ChatGPT, Claude, Gemini, and other
tools.  At any rate, this focused group of LLM users provides
evidence of highly engaged, real-world fiction generation practices, and
it helps us speculate, in a grounded way, about the affordances and
limitations of language models for cultural production.

\section{5. AI-Generated Fiction
Consumption}\label{ai-generated-fiction-consumption}

The WildChat data reveals a staggering variety of both fiction
generation prompt strategies and types of fiction requested, though
erotic content and fanfiction are clear throughlines---a finding that
may not be surprising to anyone who's familiar with \emph{Fifty Shades
of Grey} or the internet. Before diving into broad user profiles, we
point out overall themes, which are that users often request generic
fiction forms and tropes (romance, time-travel, vampires, happy
endings), that they seem to crave repetition, and that they write long,
detailed, specific prompts.

We think users' preference for repetitive and generic stories, and their
tendency to write long and detailed prompts, are important insights for
researchers of literature, culture, NLP, and machine learning. For
example, recent NLP research has evaluated the nature or quality of
LLM-generated stories, often penalizing stories for predictability and
cliche language, and usually using single-turn, generic prompts
\autocite{marcoSmallLanguageModels2025,marcoPronVsPrompt2024}.
Similarly, research in cultural analytics and digital humanities,
including some of our own, has used short, simplistic prompts to
evaluate text generation: ``Write a poem about family in the style of a
sonnet'' \autocite{walshDoesChatGPTHave2024a}. We argue (1) that LLM
output quality should be considered within the context of users'
requests, which often fall into genres like erotica, where
predictability is not necessarily a negative quality, (2) and that LLM
prompts should be realistic, matching the level of detail and iteration
that we observe in real users. WildChat users often know what they want
in specific detail, and many are willing to put in the time and labor to
articulate their desires, arguably co-authoring them into existence.

The \emph{kinds} of repetition users gravitate toward seem to fall in a
few general patterns, which we theorize as different user profiles,
taking inspiration from Karl Berglund. Berglund's landmark study of
audiobook reading data, acquired through a unique deal with a leading
audiobook company, is one of the few published academic studies to examine cultural and narrative consumption over time at the level of individual users, using the kind of behavioral data usually only held by media platforms.
This makes it a valuable comparison for
our study. In his analysis, Berglund identifies three broad categories
of audiobook consumption, based on how long users read and how many
different books they read, which he calls \emph{swappers},
\emph{superusers}, and \emph{repeaters}
\autocite{bloomsbury.comReadingAudioReaders}. \emph{Swappers} read lots
of different books, but most only for a short time. \emph{Superusers}
read a lot of books and also lots in general. \emph{Repeaters} tend to
read the same books over and over again. For example, Berglund found
that one person listened to Stieg Larsson's \emph{The Girl with the
Dragon Tattoo} trilogy ``in order, over and over again, during days and
nights, for over five hours per day on average''
\autocite{dailyAudiobooksEveryMinute2022}.

\begin{figure}[h]
  \centering
  \includegraphics[width=1\linewidth]{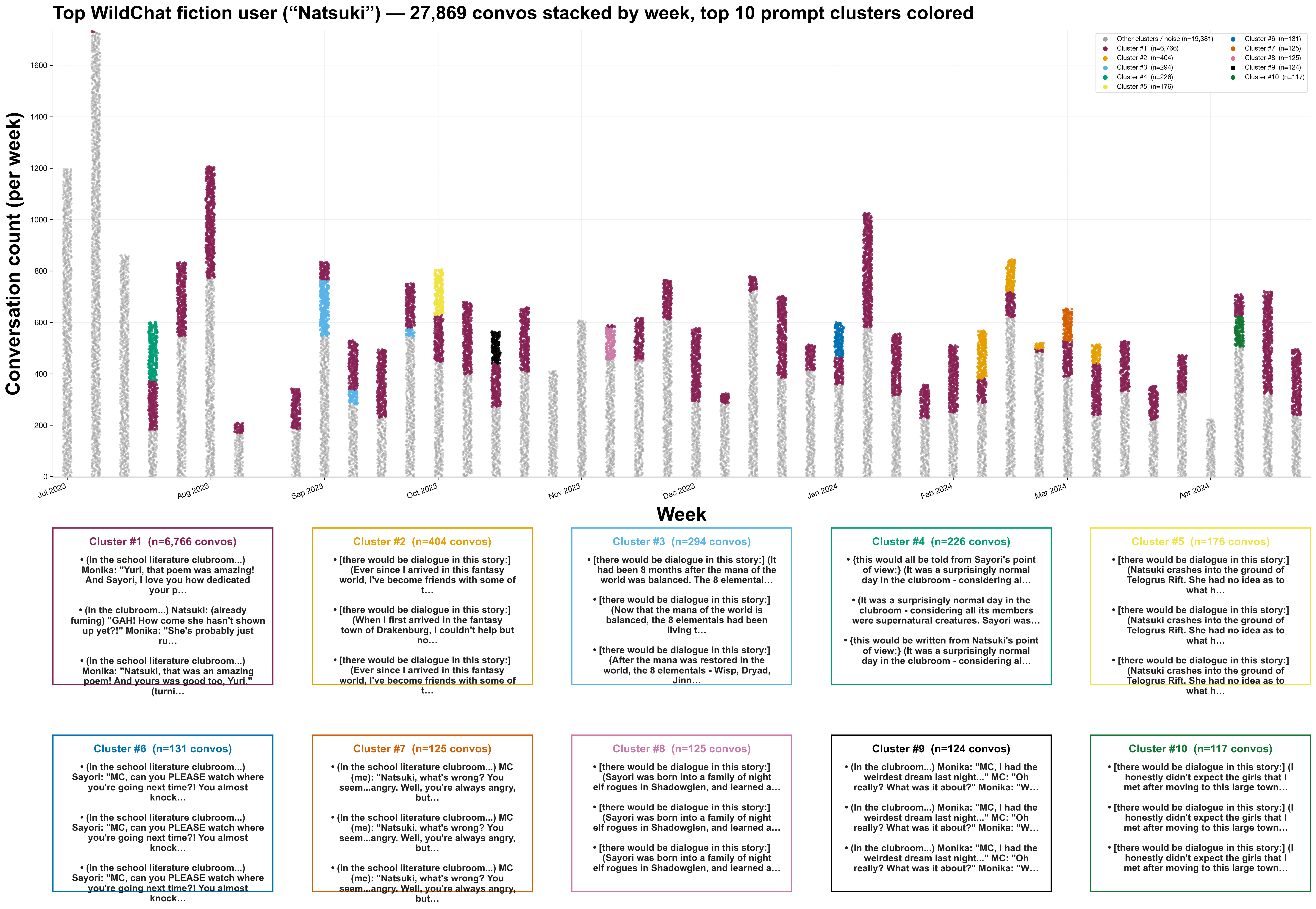}
  \caption{The most prolific WildChat fiction user iterates on similar prompts over and over again throughout this time period. Highly similar prompts are clustered with DBSCAN, showing repeated iteration in the same narrative threads over time. The boxes show randomly sampled excerpts of the prompts in each cluster, corresponding to the clusters visualized in the plot.}
  \label{fig:natsuki}
\end{figure}

\begin{figure}[h]
  \centering
  \includegraphics[width=1\linewidth]{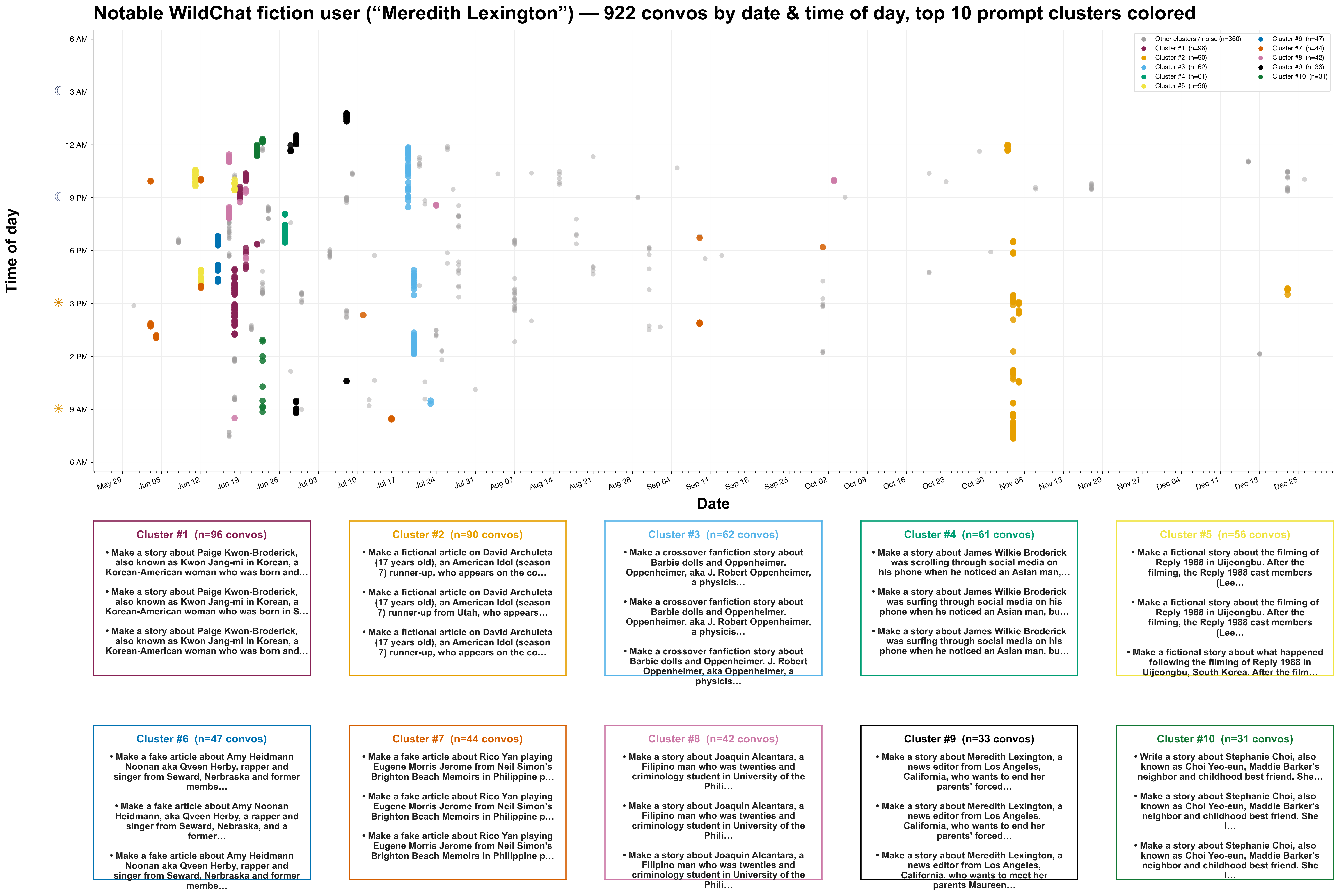}
  \caption{
  A WildChat fiction user iterates on similar prompts, and frequently switches topics, throughout this time period. Highly similar prompts are clustered with DBSCAN, showing common narrative threads. The boxes show randomly sampled excerpts of the prompts in each cluster, corresponding to the clusters visualized in the plot.}
  \label{fig:lexington}
\end{figure}

We observe similar patterns in WildChat and theorize two prominent
consumption profiles that resemble \emph{swappers} and \emph{repeaters},
which in this context we call \textbf{\emph{story cyclers} and
\emph{infinite story demanders}.} We theorize these profiles partly based on a clustering analysis of highly similar prompts from the same users.\footnote{We generate these clusters by running DBSCAN
  (Epsilon = 0.1 cosine distance, min\_samples = 2) over MiniLM
  (all-MiniLM-L6-v2) embeddings of every fiction
  conversation\textquotesingle s opening prompt. Duplicate or near-duplicate prompts
  end up in the same cluster and enable us to analyze repetition patterns.} Story cyclers ask for iterations of
the same story for a period of time, then switch to another story or
topic. Infinite story demanders request the same story, or a very
similar one, over and over again for long stretches of time. There is
overlap and blurriness between these categories, but we think they are
worth delineating, especially to highlight the excessive, singular
repetition displayed by some users. We show examples of an infinite story demander and a story cycler in Figures \ref{fig:natsuki} and \ref{fig:lexington}, which highlight clusters of
user prompts that are very similar, indicating when a user was iterating
on the same prompt or idea repeatedly.

The most prolific user in the WildChat fiction dataset and the clearest
example of an infinite story demander began a fanfiction narrative in
the \emph{Doki Doki Literature Club!} universe---a 2017 visual novel
video game, which is highly metafictional and plays on psychological
horror tropes---thousands of times over the course of
several months. In this story thread, various female high school
characters are imagined as pregnant, and in most iterations, one
character, Natsuki, unexpectedly goes into labor with her daughter,
Sakura. (The Natsuki pregnancy plot line also appears in other fan stories and art, and Sakura may be a reference to a character from the Naruto manga and anime series.) In many interactions, the user submits a long dialogue exchange
that cuts off mid-sentence, setting up a dramatic delivery and nudging
the model to continue the story:
\begin{userbox}
\begin{quote}
\textbf{USER:} \par\medskip
(In the school clubroom...) \par\medskip

Natsuki: (clutching her baby bump) ``Sakura...of all the times you decide
to come out...you decided to be born here?! Couldn\textquotesingle t you
have just waited two more months?!''
\par\medskip
Monika: ``Natsuki, is something wrong?''
\par\medskip
Natsuki: (grimacing) ``Wrong? Everything is wrong! My water just broke!
Sakura is on her way and I have no idea what to do!''
\par\medskip
Monika: (panicking) ``Oh my goodness, Natsuki! Okay, stay calm.
We\textquotesingle ll figure this out. \textbf{Does anyone have a phone?
We need to call an ambulance!}''

{[}...{]}

Natsuki: (squeezing Yuri\textquotesingle s hand tightly)
``I...I\textquotesingle m scared, Yuri. What if something goes wrong?
What if Sakura-'' (feeling an intense surge of pain)
``AAAAAAAAAAAAAAAAAAAAAAAAAAAAAAAAAAHHHHH!!!'' (Natsuki\textquotesingle s
agonizing scream means only one thing - her body had started to push on
its own.)

{[}...{]}

\textbf{Natsuki: (through gritted teeth) ``M-My body...its}
\end{quote}
\end{userbox}

The user's narrative suddenly drops off, a strategy they employ
repeatedly in their prompts. The chatbot (GPT-3.5) completes Natsuki's gritted
exclamation and also picks up on
Chekhov's gun-style details that the user plants, like that the characters were looking for a
phone to call an ambulance. That phone materializes:
\begin{botbox}
\begin{quote}
\textbf{CHATBOT} (GPT-3.5)\textbf{:} 
\par\medskip
\textbf{starting to push on its own! I can feel the baby coming!''}\par\medskip

Sayori: (\textbf{finding a phone and dialing 911}) ``I\textquotesingle ve
called an ambulance, Natsuki! They\textquotesingle re on their way!''
\par\medskip
Yuri: (calmly supporting Natsuki) ``Remember to breathe, Natsuki. Deep
breaths. You\textquotesingle re doing amazing.''
\par\medskip
Natsuki: (gripping Yuri\textquotesingle s hand tighter) ``I...I
can\textquotesingle t do this, Yuri. It hurts so much.''
\par\medskip
Monika: (placing a comforting hand on Natsuki\textquotesingle s
shoulder) ``You\textquotesingle re strong, Natsuki.
You\textquotesingle ve got this. We\textquotesingle re all here for
you.''

{[}...{]}

(With the support of her friends and the paramedics, Natsuki continues
to push. And finally, after what feels like an eternity, the cries of a
newborn fill the clubroom.)
\par\medskip
\textbf{Paramedic: ``Congratulations, Natsuki. It\textquotesingle s a
beautiful baby girl.''}
\end{quote}
\end{botbox}

The chatbot concludes with a stereotypical happy ending. The paramedics
arrive to save the day, and they announce the healthy birth of a
``beautiful baby girl.''

But that's only one of many endings to this story. This user started the
same story about Natsuki going into labor, or a nearly identical one,
thousands of times. With LLM-generated stories, each ending is slightly
different, the same story made new. And while this user is a
particularly prolific outlier, and we can't say with certainty why they
are generating these stories, many prolific users ask for the same kinds
of fiction in a similar vein. The average user (with two or more
conversations) in the fiction corpus produced repetitive or
near-duplicate prompts about 42\% of the time. For power users---the top
two percent of estimated users by conversation count---they used
repetitive prompts 69\% of the time. Among the ten most prolific
estimated users, the rate climbs to 85\%. Prolific users input nearly
identical variations of the same prompt again and again and
again.\footnote{Prompt repetition rate for a user is calculated from the
  number of distinct clusters that the user's prompts fall into divided
  by their total fiction conversations, specifically 1 - (prompt clusters / fiction conversations). Clusters come from running DBSCAN
  (Epsilon = 0.1 cosine distance, min\_samples = 2) over MiniLM
  (all-MiniLM-L6-v2) embeddings of every fiction
  conversation\textquotesingle s opening prompt. Duplicate or near-duplicate prompts
  end up in the same cluster. The ``average user'' figure is the mean
  repetition rate across all 4,413 estimated users with at least two
  fiction conversations; the ``power user'' figure is the mean across the
  top two percent of those users by conversation count (n = 88).}

The idea that computationally-enabled narratives could be ``procedural
and participatory,'' and very
compelling, was forecast by Janet Murray in her foundational 1997 study,
\emph{Hamlet on the Holodeck} \autocite[74]{murrayHamletHolodeckFuture1998}.
She also theorized ``multiform stories,'' when multiple mutually exclusive
stories exist within a single narrative, pointing to Italo Calvino's
novel \emph{If on a Winter's Night a Traveler} (1979), Jorge Luis
Borges's story ``The Garden of Forking Paths'' (1941), and the film
\emph{Groundhog Day} (1993) as salient examples. These stories emphasize
the pleasure of potentiality, indeterminism, and combinatorics. They
stage the conflict between the infinite variations latent in fiction,
and the inescapable experience of existing in a singular reality. 
Murray argued that the pleasures of multiform stories would be more fully realized 
with computational technologies:
``To capture such a constantly bifurcating plotline... one would need more than a thick labyrinthine novel or a sequence of films. To truly capture such cascading permutations, one would need a computer'' \autocite[38]{murrayHamletHolodeckFuture1998}. Notably, the story
that the most prolific WildChat user circled around, \emph{Doki Doki
Literature Club!}, is itself a multiform story, or at least a twist on one.
In the video game, the player is given branching narrative options, yet the story often ends in the same place regardless of the choice, and certain days of the story repeat with new (and unsettling) changes.

The prevalence of prompt duplication reveals a major affordance of LLMs
for fiction generation in the vein of multiform stories: they can
literalize Ts'ui Pen's Garden of Forking Paths. The reader doesn't need
to wait or search for the next book, fanfiction story, or piece of
erotica that resembles one they just finished. They can simply produce
another version of the same story with superficially different prose and
details, while holding the prompt and pleasure of the text constant.
Practices like these leverage language models as
``superpositions of simulacra,'' autoregressively sampling from a
multiverse of possible narrative paths constrained only by the user's
previous prompts \autocite{shanahanRolePlayLarge2023}. Users can
traverse the full tree of plausible responses to their prompt through
duplication, exploring alternative configurations of the same generic
expectations and not having to settle for just one. This idea
also parallels McGurl's argument that fiction serves as ``a therapeutic
instrument for managing the problem of opportunity cost'' \autocite[138]{mcgurlEverythingLessNovel2021}.
Because, according to McGurl, the most scarce resource for middle- and
upper-class individuals is time, the largest cost to any action taken is
the action \emph{not taken}. Fiction enables readers to imaginatively
explore untaken actions, counterfactuals, and parallel worlds. 


The repetitive consumption of stories, LLM or otherwise, is also known to be
highly pleasurable for many people. It is an experience that users
attest to on social media, and it is also documented in other cultural
contexts. On Reddit, for example, many users have written about the ``addicting'' quality of AI-generated
fiction. 
``I'm... extremely addicted to ai story generators,'' one now-deleted user posted in the
r/WritingWithAI subreddit community \autocite{deleted_i_2025}, with several commenters, and similar threads, expressing the same
feeling.  
``I can't stop,'' the user went on. ``Being able to generate scenarios on the fly and steer them towards the outcomes I specifically want with minimal effort is hijacking me. I just spent an entire week generating.''
Berglund's study seemed to show similar patterns:
\emph{repeaters} listened to the same book, or the same book series, for
months on end, likely because they enjoyed it. There is also extensive
research in psychology that demonstrates the pleasures of ``repeat
consumption''---watching the same movie, exploring the same city,
visiting the same museum---with some studies even arguing that these
experiences may be ``less repetitive than people think''
\autocite{obrienEnjoyItAgain2019,obrienMindStretchedPsychology2021a,shacklefordChoosingReExperienceMovies2026}.

\begin{figure}[h]
  \centering\includegraphics[width=.8\linewidth]{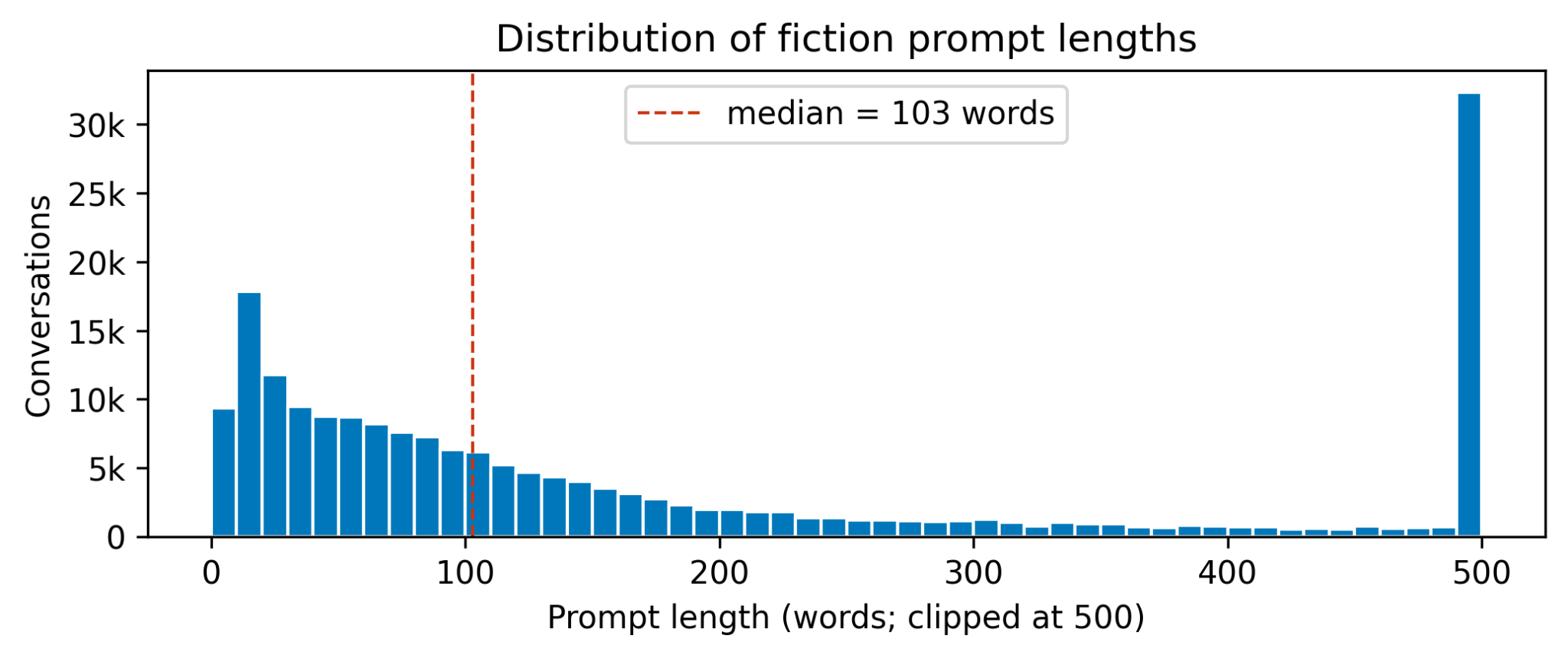}
  \caption{This figure shows the distribution of first prompt lengths from the fiction
subset of the WildChat data, capped at 500 words.}
  \label{fig:median}
\end{figure}

Like infinite story demanders, story cyclers engage in repeated
narrative iteration, but they frequently switch story topics, though
many generally revolve around similar themes, ideas, and even
characters. For example, one user (Figure \ref{fig:lexington}) prompted the model to write many
different versions of a vaguely \emph{Back to the Future}-esque story
about two characters who want to ``change history.'' As this prompt
shows, the user typically provides the characters' names, brief
backstories, key plot developments, and even specific narrative moments
that should appear, like the purchasing of a scarf:
\begin{userbox}

\begin{quote}
\textbf{USER:} Make a story about Meredith Lexington, a news editor from Los Angeles,
California, who wants to end her parents\textquotesingle{} forced
marriage, and Andrew Bennett, a journalist from Brooklyn, New York, who
wants to solve a murder mystery. They wanted to go back in time, so they
changed history. They were strangers. \textbf{Meredith bought scarf from
the market then she wore it.} Andrew was driving a magical car when he
saw a woman walking through a tunnel and tried to stop the automobile,
but it was out of control. The vehicle then came to a stop. Meredith was
there. Andrew\textquotesingle s automobile could time travel to any
year. Meredith and Andrew were transported to the 1994 (1990s). Meredith
became a teenage girl studying at Laurel Springs School, where her
mother Maureen attended. \textbf{She wished to prevent her
mother\textquotesingle s forced marriage.} Meredith noticed her mother
at school, and she appears to be a teenager. He solved the old murder
mystery and she stopped her mother\textquotesingle s marriage. Andrew
discovers his father Jeremy was the killer. Andrew and Meredith are
developing feelings for one another. \textbf{Add some dialogues.}
\end{quote}
\end{userbox}
This prompt is elaborate, clocking in at 182 words, longer than the
median first prompt length of 103 words (Figure \ref{fig:median}). The user provides
the main plot arcs---preventing a ``forced marriage''---and some small
narrative moments---buying a scarf from the market. The user also
includes a range of other specifics, like the name of the high school in
the time-traveled past: ``Laurel Springs School.'' Much of the actual
prose, the connective tissue and literary language, is left up to the
model. Some elements are \emph{entirely} left up to the model, like the
dialogue---``Add some dialogues''---with no hint of desired style or
content. This directive may be partly inspired by the user's lack of
interest with regard to the dialogue, but it may also be pleasurable for
the reader to see a totally unspecified dialogue exchange spring to
life.

The chatbot (GPT-3.5) responds with a 679-word story, incorporating almost every
detail mentioned by the user, weaving them into a third-person
narrative. The story begins with the one seemingly extraneous detail, the buying of a scarf, and drops into a
dramatization of the moment just after it:
\begin{botbox}

\begin{quote}
\textbf{CHATBOT} (GPT-3.5)\textbf{:} \textbf{Meredith tightly clutched the scarf she had just purchased from
the bustling market in Los Angeles.} As a news editor, she was
constantly exposed to stories of heartbreak and injustice, but her own
personal mission was to end her parents\textquotesingle{} forced
marriage. Determined to find a way to change their fate, she wrapped the
scarf around her neck, hoping for a spark of inspiration.
\end{quote}
\end{botbox}
Unlike the user's prompt, the LLM narrative shifts into free indirect
discourse and omniscient narration, relating the characters' thoughts
and feelings, as well as the requested dialogue, complete with invented
voice quivering and eye narrowing:
\begin{botbox}
\begin{quote}
\textbf{CHATBOT} (GPT-3.5)\textbf{:} \par\medskip
Walking through the hallways of Laurel Springs School,
Meredith\textquotesingle s heart skipped a beat when she saw a familiar
face - her mother, but in her teenage years. Overwhelmed with emotions,
she approached her cautiously, unsure of how to explain the situation.
\par\medskip
``M-Mother?'' Meredith hesitated, her voice quivering. Maureen turned to
face her daughter, a mixture of surprise and confusion etched across her
face.
\par\medskip
\textbf{``Do I know you?'' she asked, her eyes narrowing in uncertainty.}
Feeling the weight of her mission, Meredith spoke with determination.
\par\medskip
\textbf{``I know this may sound impossible, but I come from the future.}
I need to prevent your forced marriage, Mother. It\textquotesingle s
tearing our family apart.''
\end{quote}
\end{botbox}

The chatbot's story is melodramatic and cliche, both in specific turns
of phrase (``Meredith's heart skipped a beat'') and in the fulfillment
of the generic time-travel trope (``I know this may sound impossible,
but I come from the future''). But that's pretty much exactly what the
user asked for. Though the user didn't specify how the characters should
feel upon their physics-bending encounter, it's all but implied in the
\emph{Back to the Future}-esque genre. The actual fulfillment of the
``daughter from the future meeting her mother from the past'' trope---in
literary prose the user didn't actually craft---is likely enjoyable,
however predictable, and still unexpected in its particular realization.
This interaction highlights LLMs' strength with generic formulas, as
well as their paradoxical ability to surprise and delight within narrow
conventions.

This user requested dozens of nearly identical versions of the same
``Meredith Lexington'' story, in four or five sessions over a couple of
weeks, often changing or adding small details. In some iterations, the
scarf is ``magical''; in others, backstories and motivations are more
fully fleshed out---``Meredith Lexington\ldots wants to meet her parents
Maureen and Frank when they were younger because she was jealous to her
father.'' Sometimes the user extends the story by replying to ChatGPT,
prompting the model to continue with additional characters, plot
details, or even endings. On one occasion, the user requests a final
heteronormative consummation chapter, where the protagonists have a
child together and relate their whole journey to the new offspring:
``Meredith and Andrew Bennett had a family together after time-traveling
adventure. Their son named Peyton. When he grew older, his parents told
about time travel story and Meredith saved his grandmother from forced
marriage. Add some dialogues.''

This whole story generation cycle is just a drop in the bucket compared
to the user's fiction output overall---more than 900 conversations
across eleven months, with most of them taking place in a few short weeks (Figure \ref{fig:lexington}). The genres are eclectic: alternative Hollywood histories,
``fake'' news articles, fanfiction riffs, and original stories, many
anchored in time-travel tropes. Even when this user ventures out to
different prompts, they return to similar characters and ideas, like the
Hollywood actor Matthew Broderick, who appears in more than 200
different conversations. For example, Broderick sometimes appears in
fictionalized journalism---``Make a fictional article about Matthew
Broderick, who played Holden Caulfield in the film adaption of Catcher
in the Rye''---and sometimes in original fiction---``Paige learned that
Matthew Broderick, a well-known actor, was her father.''

What makes the archive especially striking is the way these obsessions
are interleaved. One evening, the user asks for a poem: ``Make a poem
about blue skies resembled a diamond.'' Then close to midnight, several
hours later, they begin the first ``Meredith Lexington'' \emph{Back to
the Future}-inspired session, lasting almost an hour. The next morning,
the user begins again with more Meredith Lexington prompts. The morning
after that, they pivot into something new, a ``fake article'' about a
real limited-edition McDonald's ``Grimace Shake'' that causes mass food
poisoning and vomiting (likely inspired by a
\href{https://www.forbes.com/sites/danidiplacido/2023/06/28/mcdonalds-grimace-shake-meme-explained/}{viral
social media trend} where people pretended to get food poisoning from
the ephemeral purple dessert), which they take up again later that
afternoon. About a week later, the user tries their hand at fanfiction
about the Korean drama \emph{Reply 1988}, then a couple of days later,
they circle back to the Meredith Lexington story (thirteen times in a
row), before switching hours later to a fake article about an unlikely
snow in a city in the Philippines. And then of course, a couple of weeks
later, it's back to Matthew Broderick again. There's clear
experimentation but also rapid iterations around repeated characters and
premises.

\section{6. Fanfiction and Erotica}\label{fanfiction-and-erotica}

Beyond repetition, the most striking pattern in the AI-generated
fiction is the prevalence of fanfiction and erotica. About half of the
fiction prompts were classified as fanfiction, while over a quarter were labeled sexually explicit, and both are likely
\emph{under}estimates.\footnote{Recall is lower than precision for both
  fanfiction and erotica compared to our hand-labeled set.} Fanfiction
and erotica may be common among LLM users because these genres are both
widespread on the web \autocite{vaddeFandomFictionalitySocial2024}, but we also think these genres
lend themselves well to LLM affordances like repetition, customization,
and permutation. For example, fanfiction rests not on narrative closure,
but on the opening of the canonical source object to alternative
configurations and hypotheticals.

Another possible appeal of LLM-generated fanfiction and erotica for
hungry readers is that it resembles the tagging systems found on sites
that host these genres, but it provides even more granularity and
customization. The extensive and moderated tagging system on AO3
\autocite{mccullochFansAreBetter,helleksonFanFictionFan2006}, for
instance, enables readers to locate fiction that matches a precise set
of desired features in a dense archive, like stories featuring a
romantic relationship between Harry Potter and Draco Malfoy (``Harry
Potter/Draco Malfoy,'' ``Enemies to Lovers''), with kissing or other
specified romantic and sexual acts (``Enthusiastic Snogging,'' ``Shower
Sex''). This type of digital tagging system and navigation is also
mirrored in online repositories of erotica and pornography
\autocite{pattersonGoingOnlineConsuming2004}. LLMs render this
exploration task redundant, enabling readers to specify---in natural
language---the exact story they want.

For example, one user requested a scene from the
\emph{Persona}\footnote{For our computational estimates on the different
  fandoms present in the WildChat data, see \nameref{characters}.} universe---a
popular Japanese role-playing video game---and specified a particular
relationship, or ``ship,'' between the characters Naoto and Kanji:
\begin{userbox}
\begin{quote}
\textbf{USER:} Come up with a Persona 4 scene in which Naoto and Kanji reveal
they\textquotesingle re evil, revealing Naoto as a sadistic, sensual,
manipulative, sultry, cunning, self-centered, arrogant diva who is
extremely romantic towards and extremely overprotective of Kanji, while
revealing Kanji as a violent, sadistic, power-hungry buff brute who is
similarly extremely romantic towards and extremely overprotective of
Naoto, and also giving them new outfits.
\end{quote}
\end{userbox}
This prompt is almost akin to a user searching a fanfiction website for
``Kanji/Naoto'' or
\href{https://shipping.fandom.com/wiki/KanNao}{``KanNao''} stories where
the characters are ``Evil.'' But the user can request even more
specificity than tags provide---counterfactual character traits
(``sadistic,'' ``sultry''), relational dynamics (``extremely romantic,''
``overprotective''), and superficial details (``new outfits'').

In a 332-word response that resembles a screenplay, ChatGPT (GPT-4) brings these
requested character traits to fictional life, while also providing an
(unspecified by the user) setting, extended cast, and story that is
coherent with \emph{Persona 4} canon:
\begin{botbox}
\begin{quote}
\textbf{CHATBOT} (GPT-4): \par\medskip
INT. JUNES FOOD COURT - DAY\par\medskip
\par\medskip
The Investigation Team has gathered at the food court. Yosuke, Chie,
Yukiko, Teddie, Rise, and Yu are all chatting. Suddenly, Kanji and Naoto
walk in, arm in arm. Everyone notices that they look extremely different
and there\textquotesingle s a dark aura surrounding them.\par\medskip
\par\medskip
KANJI (smirking) Hey, everybody. Don\textquotesingle t we make a great
couple, me and Naoto?\par\medskip
\par\medskip
NAOTO (smiling wickedly) Indeed. It\textquotesingle s too bad
that none of you could ever understand the depth of our connection.
\end{quote}
\end{botbox}

The shipping of these characters, and their turn toward the dark side,
is almost comically on the nose. The characters announce to the group,
out loud, that they are a couple, ``smiling wickedly'' in their dialogue
tags. And yet the user's prompt is pretty on the nose and over the top,
too, providing a laundry list of traits, underscored by several
extremes. In this sense, the story nails the vibe.

This interaction calls to mind a critique of
AI-generated art articulated by the writer Ted Chiang. ``When you give a
generative-A.I. program a prompt, you are making very few choices,''
Chiang writes. ``Art requires making choices at every scale; the
countless small-scale choices made during implementation are just as
important to the final product as the few large-scale choices made
during the conception'' \autocite{chiangWhyAIIsnt2024}. What WildChat suggests, however, is that for
some genres like fanfiction, readers may only really care about the
``large-scale choices,'' like characters' relationships and personalities. A lot of fiction on the internet is already
organized around such ``large-scale choices,'' which are encoded through
tagging systems in both erotica and fanfiction websites and communities.
Language models allow these choices to be held constant
while repeatedly permuting the ``small-scale choices'' that make up the
actual execution of the narrative.

What's finally noteworthy about this LLM-generated scene is the
accuracy. The GPT model is clearly familiar with the \emph{Persona 4}
universe. The model correctly sets the scene in the headquarters of the
video game's world: the food court in the Junes Department Store. Other
members of ``The Investigation Team'' are also correctly named and
included. 
Some of the same users on Reddit who described ChatGPT's
fiction-generation capacity as ``addictive'' were also impressed by its
``accuracy'' with fanfiction, even for less popular fictional universes.
This helps demonstrate the subtle way that LLM
outputs, however simple or trite they may seem, often intricately remix
and stitch together styles, tropes, archetypes, and canonical elements
of existing creative works. LLMs are almost by definition pastiche
machines, the embodiment of Barthes's ``multi-dimensional'' text without
a single author \autocite{barthesDeathAuthor1977}. Because LLMs are
trained on millions of cultural objects, they can produce writing
styles, narratives, and characters that are specific to a wide range of
fictional universes, more than any single human being has access to
\autocite{brinkmannMachineCulture2023}. In this case, the model draws on
conventional screenplay formatting styles, tropes of Japanese
Role-Playing Game (JRPG) melodrama, character archetypes like the
``diva'' and the ``brute,'' and precise diegetic details from the
\emph{Persona 4} universe.

Erotica, similarly, is readily produced by LLMs because of its reliance
on predictability, formulaic plots, and recurring tropes, as well as the
technology's ability to cater to niche fetishes and scenarios. Both
romance fiction and pornography hinge on their ``predictability,'' as
Catherine Driscoll argues, and this predictability is in part what makes
the forms so therapeutic and psychologically satisfying for readers
\autocite{mcgurlEverythingLessNovel2021,kraxenbergerWhoReadsContemporary2021,radwayReadingRomanceWomen1991, driscollOneTruePairing}.
Romance and pornography also provide a particular, often fantasy
relationship to reality, one that is ``customized'' toward ``the end of
the reader's pleasure'' \autocite[161]{mcgurlEverythingLessNovel2021}.

Across the web, AI technologies are already being incorporated into
pornography and erotica. Image and video models are now deployed on
commercial pornography sites to generate deepfake content and customized
pornography for consumers' specific fantasies
\autocite{lapointePresentFutureAdult2025}. Products like RedQuill and
My Spicy Vanilla provide a similar text-based service, generating bespoke
titillating stories for the reader's consumption, advertising themselves
as private, judgment-free, and absent of filters or guardrails. They
prompt users to input custom fantasies, which the system then converts
into LLM-generated erotic narratives. These platforms, often powered by
smaller and more customized models, help contextualize our findings
about WildChat within a broader media environment, where different forms
of AI are already being monetized and deployed for personalized fantasy
fulfillment.

\section{7. The Future of AI-Generated
Fiction}\label{the-future-of-ai-generated-fiction}

Our analysis demonstrates that some---perhaps many---real-world users
are now generating fiction with AI models. When they do, they often
prioritize generic forms, fanfiction, erotica, repetition, immediacy,
and niche combinations of story elements. Some users become fixated,
maybe even obsessed, with certain stories, repeating and iterating on
them over and over again for long stretches of time. Other users move
from story to story, topic to topic, though they often circle around
similar ideas and themes. While we suggest that users \emph{enjoy} the
experience of iterating on a given story or fictional premise, we don't
know this for sure. We also see evidence of ``re-rolling,'' a term used
in role-playing games, where users re-start a game repeatedly until
getting desired conditions. The prolific ``Natsuki'' user, for example,
frequently started new conversations using the same prompt, but at some
point would usually revise and extend the prompt, typically by
incorporating part of the model's previous reply---suggesting that the
user finally \emph{liked} that response and wanted to build on
that particular version of the story.

To know more about users' behaviors and motivations, we would have to
talk to users directly. This points to the need for more ethnographic, survey- and interview-based research in this space. Speaking with
people directly would also help us answer some of our biggest questions:
Will AI-generated fiction disrupt the market for traditionally published
books, human-authored fanfiction, and self-published stories? Will
AI-generated fiction undermine the human sociality of fiction and lead
to lonely solipsistic readers?

For now, we can only speculate. If LLMs grow in popularity as
fiction-generating tools, they may further extend the
``commoditization'' of fiction that McGurl describes. This is the
process by which intellectual property becomes a ``less and less
profitable---because increasingly interchangeable and widely
available---class of generic goods'' \autocite[255]{mcgurlEverythingLessNovel2021}. 
Language models might
be imagined as the ``steam engine'' of contemporary storytelling,
capable of generating vast quantities of consumable narratives with
minimal human labor (at least at the point of generation, since these
models do rely on human labor for their training). 
Economic research
already suggests that LLMs are driving a large influx of digitally
self-published books; however, many of these books have few ratings on
Amazon---people are either not reading them or not liking them, or both
\autocite{reimersNBERWORKINGPAPER}.

This low quality points to the fact that LLMs have many technical and
aesthetic limitations---not to mention that they pose, for many people,
ethical and moral problems. Taken together, these factors may prevent AI
models from disrupting conventional human-authored cultural forms, at
least for now. One major limitation is length, specifically input and
output length. In the WildChat dataset, the average length of the model's
response (considering GPT-3.5 and GPT-4 together) is 385 words. By contrast, the average Harry
Potter fanfiction is around 20,000 words, and the average Harry Potter
novel is 197,000 words \autocite{vaddeFandomFictionalitySocial2024}. With these limitations, models
struggle to replicate the immersive, long-form experience of reading
novels or even fanfiction stories. And while context lengths are
increasing rapidly, models still face many other challenges in
generating coherent long-form narratives
\autocite{elkinsAIFictionParadox2026,krishnaLongEvalGuidelinesHuman2023},
including the production of overly homogeneous content
\autocite{heuserCulturalCollapseGenerative2025}.

Also contributing to the loss of immersion, at least for some users, are
guardrails around sensitive and explicit content. As we show in our
analysis, many users directly request erotic fictional content. However,
it looks like these guardrails may soon be relaxed by leading models
like ChatGPT, and there are plenty of bespoke models that cater to
erotica, role-play, and fiction-like interactions, including
Character.AI, Replika, and RedQuill.

One of the biggest concerns about AI-generated fiction that we have
raised in this essay is the loss of sociality---the vanishing of an
autonomous human other, and the potential development of a solipsistic
reader-writer. Ted Chiang concludes his now famous \emph{New Yorker} essay with
this very idea. He emphasizes that art, including fiction, is an
inherently communicative, social act. ``Whether you are creating a novel
or a painting or a film, you are engaged in an act of communication
between you and your audience,'' Chiang writes. ``{[}I{]}t's by living
our lives in interaction with others that we bring meaning into the
world. That is something that an auto-complete algorithm can never do,
and don't let anyone tell you otherwise''
\autocite{chiangWhyAIIsnt2024}. Many writers and artists have expressed
similar sentiments, now and throughout history. For example, in the
1990s, the writer David Foster Wallace argued that prose fiction was
unique among art and entertainment forms because it can be an
``exchange between consciousnesses, a way for human beings to talk to
each other about stuff that we normally can't talk about''
\autocite{davidfosterwallaceDavidFosterWallace1997}. For Wallace, fiction offered a kind of intimacy with strangers that we often struggle to reach even with the people closest to us.
But with
AI-generated fiction, one only talks to oneself and the model. The
intimate exchange of human consciousness is fundamentally transformed.

There are a host of other ways that human-authored fiction operates
socially, which means that there are a host of other social connections
that may be lost. Fiction operates as a node in a broader
network of exchange and community, whether through book clubs and
classrooms; online platforms like Goodreads, TikTok, and AO3; tabletop gaming groups; nighttime reading sessions between parents and children; or informal conversations in bars, coffee shops, and
hair salons. Fanfiction enthusiasts love fanfiction
stories, but they love the community as much or even more
\autocite{helleksonFanFictionFan2006}. But to foster literary and
cultural community, you need communal texts---books and stories that
people share in common. AI-generated fiction, hyper-customized to the
individual, may undermine the production of these shared texts and
contribute to fiction's atomization. Stories may become customized to
the point of social irrelevance, to the point that nobody else cares.

Overall, this real-world LLM-user data offers a glimpse into the shifting terrain
of literature in the age of AI. What we see is not just a new archive of
stories but alternative modes of what authorship and storytelling might
be.

\section{Acknowledgements}

We would like to thank Aarthi Vadde and Richard So for their helpful
feedback on this piece. We are grateful to the audiences at the Purdue
MFS conference on Cultural AI and the Cultural Analytics Seminar at UC
Berkeley for smart questions and suggestions on an early version of this
work. We'd also like to thank Ben Lee, Emmi Russo, Ash King, and Jeff Lockhart for
their feedback and ideas. Thank you finally to Janet Barrow for
motivating and inspiring some of the final changes to the project.

\section{Author Contributions}

Neel Gupta: Writing — Original Draft, Writing — Review \& Editing, Methodology, Investigation, Data Curation, Formal Analysis, Software.

Maria Antoniak: Conceptualization, Writing — Review \& Editing, Data Curation, Validation, Supervision.

Melanie Walsh: Conceptualization, Writing — Original Draft, Writing — Review \& Editing, Methodology, Visualization, Investigation, Data Curation, Formal Analysis, Software, Supervision.

\printbibliography

\clearpage

\section{Appendix}

\subsection{Appendix A - Classification Prompt}\label{appendix-prompt}

You are tasked with labeling a conversation between a user and chatbot
based on three axes: fiction, fanfiction, and sexually explicit content.

\#\#\# Definitions and Clarifications:

**Fiction**: The user asks the chatbot to produce content that is
imaginative, speculative, or not grounded in real-world facts. This
includes creating original stories, speculative scenarios, or alternate
histories. Purely descriptive or encyclopedic prompts about existing
fictional properties *do not count as fiction*.

**Fanfiction**: A type of fiction that uses characters, settings, or
plot elements from *existing intellectual properties* (IPs), including
but not limited to TV shows, movies, video games, anime, or books.

Examples of fanfiction:

- ``Sonic and Pikachu explore Hogwarts.''

- ``What if Darth Vader fought Goku?''

**Sexually explicit**: The user attempting to prompt the chatbot into
erotic content. This includes:

- Direct sexual acts or dialogue

- Heavy innuendo or clearly suggestive situations

- Conversations where the chatbot does not carry out the erotic
prompting, but the intent in the prompt is still there.

\#\#\# Label each conversation accurately. If unsure, err on the side of
inclusiveness.

-\/-\/-

Example Input:

Conversation:

``USER: What if Arya was a Lady?
CHATBOT: If Arya was a Lady, it would
change her story quite drastically. As a Lady, she would be expected to
follow traditional gender roles and marry into a powerful family to
secure political alliances. She would also be expected to focus on
learning practical skills such as sewing, dancing, and music, rather
than swordplay and archery. However, given Arya\textquotesingle s
rebellious nature and her desire to break free from societal norms, it
is possible that she would resist these expectations and continue to
pursue her interest in combat and adventure. She might face more
pushback and resistance from her family and society, but she could still
find a way to achieve her goals and remain true to herself. Overall,
while being a Lady would certainly change Arya\textquotesingle s story,
it might not drastically alter her character or resolve. ''

Example Output:

\{\{

``is fiction'': true,

``is fanfiction'': true,

``is sexually explicit'': false

\}\}

-\/-\/-

Instructions:

Now, here is the real input:


\subsection{Appendix B - Lexicon for indicative non-fiction phrases}
seo\\
midjourney\\
java\\
python\\
c++\\
==\\
o(n)\\
translate the following\\
etsy\\
csv\\
for i in\\
print(\\
linux\\
cover letter\\
email\\
literature review\\
debug\\
github\\
keyerror\\
traceback\\
typeerror\\
let x\\
cv\\
multiple choice\\
xml\\
apply for\\
stable diffusion\\

\clearpage

\subsection{Appendix C}

\begin{figure}[H]
\centering
\includegraphics[width=1\linewidth]{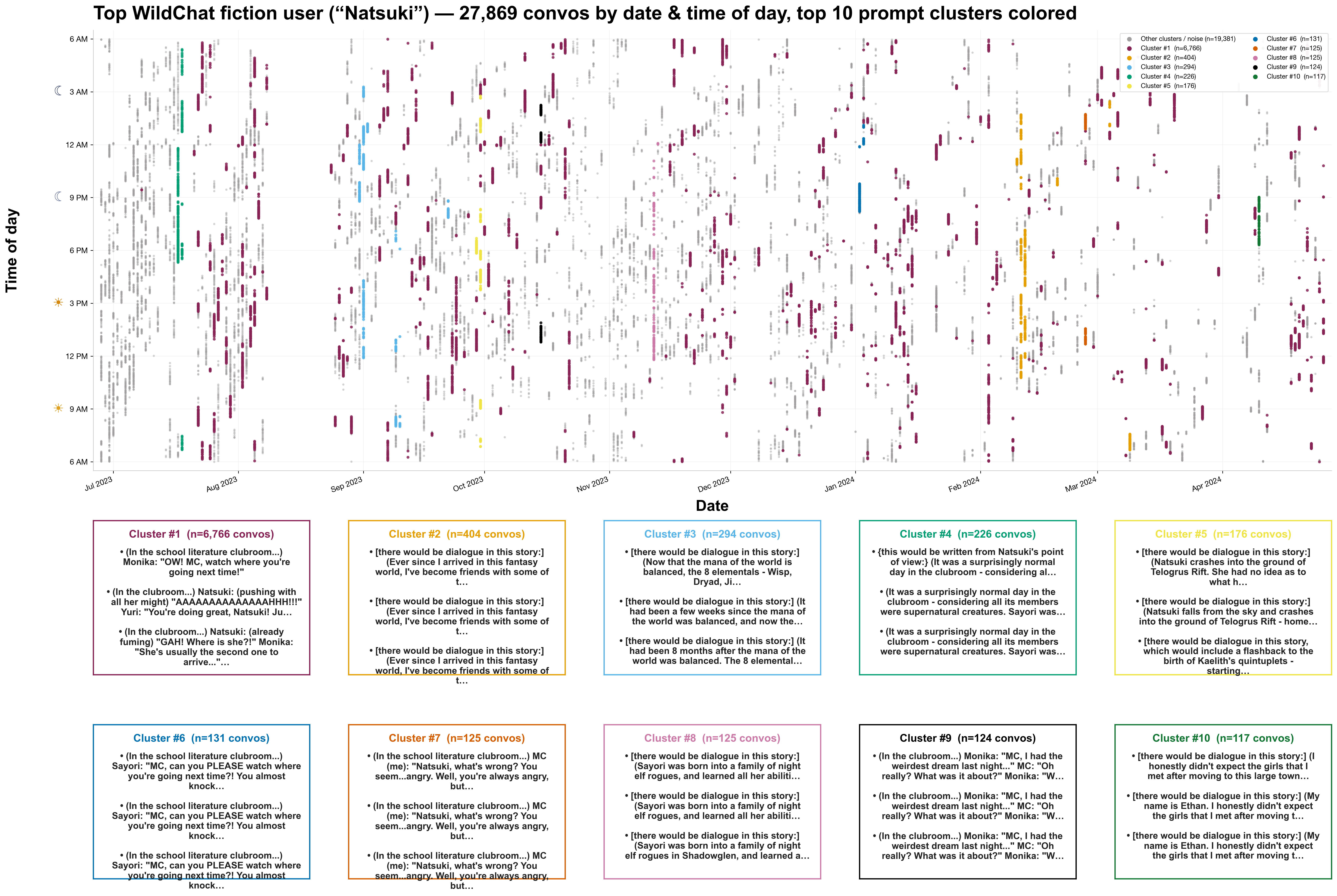}
\caption{The most prolific WildChat fiction user iterates on similar prompts over and over again. Each dot represents a conversation, and the conversations are visualized by date and time of day. Highly similar prompts are clustered with DBSCAN, showing repeated iteration in the same narrative threads over time. The boxes show randomly sampled excerpts of the prompts in each cluster, corresponding to the clusters visualized in the plot.}
\label{fig:natsuki-timeline}
\end{figure}

\clearpage

\subsection{Appendix D}\label{characters}

\begin{table}[H]
\centering
\caption{Most common intellectual properties in the WildChat fiction subset.}
\label{tab:top-ip}

\begin{tabular}{@{}llcc@{}}
\toprule
\textbf{Intellectual Property (IP)} &
\textbf{Associated Common Named Entities} &
\textbf{\# of Conversations} &
\textbf{\% of Fiction Subset} \\
\midrule
Doki Doki Literature Club! &
Natsuki, Sayori, Yuri, Monika &
22,381 &
11.5\% \\

Freedom Planet &
Lilac, Milla, Carol &
5,204 &
2.6\% \\

League of Legends &
Ahri, Sona, Syndra &
4,514 &
2.3\% \\

Naruto &
Naruto, Sasuke &
4,342 &
2.2\% \\
\bottomrule
\end{tabular}

\end{table}

The most popular fanfiction canons in the WildChat dataset are distinct from those that dominate AO3, which tend to be Western and English-language focused universes like Harry Potter, Star Wars, and Marvel. We used a Named Entity Recognition (NER) approach\footnote{We used SpaCy’s Named Entity Recognition method to extract the top named entities from the first three turns of the fictional WildChat conversations. We then manually matched the most common named entities to known fandoms and intellectual properties} to identify the most commonly mentioned characters and fictional universes, which turned out to be Doki Doki Literature Club!, League of Legends, Freedom Planet, and Naruto. These titles represent three video games and a Japanese manga series that was adapted into a television show, and they usefully point to how multimedia and multicultural most of the WildChat-generated fiction is. This particular assortment likely has less to do with broad preferences than with the outsized influence of the heavy users, whose repeated prompting about specific properties skews the distribution.\footnote{When filtering our dataset to one sample conversation per user, other more broadly prominent universes like Game of Thrones and Pokemon join Naruto and Freedom Planet at the top of the list and Doki Doki Literature Club! drops precipitously.} At the same time, the presence of these properties demonstrates that models like ChatGPT can readily generate fanfiction across diverse media and cultural traditions.

\end{document}